\documentclass[letterpaper]{article} 
\usepackage{aaai24}  
\usepackage{times}  
\usepackage{helvet}  
\usepackage{courier}  
\usepackage[hyphens]{url}  
\usepackage{graphicx} 
\usepackage{natbib}  
\usepackage{caption} 
\usepackage{algorithm}
\usepackage{algorithmic}
\usepackage{bm}
\usepackage{overpic}
\usepackage{amssymb}
\usepackage{amsmath}
\usepackage{booktabs}
\usepackage{color}
\usepackage{xcolor}
\usepackage{multirow}
\usepackage{makecell} 
\definecolor{shadecolor}{RGB}{150,150,150}
\definecolor{Gray}{gray}{0.5}
\definecolor{Highlight}{HTML}{39b54a}
\usepackage{xcolor}
\usepackage{soul}
\usepackage[misc]{ifsym}
\definecolor{frenchblue}{rgb}{0.0, 0.45, 0.73}

\usepackage{newfloat}
\usepackage{listings}
\DeclareCaptionStyle{ruled}{labelfont=normalfont,labelsep=colon,strut=off} 
\floatstyle{ruled}
\newfloat{listing}{tb}{lst}{}
\floatname{listing}{Listing}
\title{SAM-PARSER: Fine-tuning SAM Efficiently by Parameter Space Reconstruction}
\author{
    Zelin Peng\equalcontrib,
    Zhengqin Xu\equalcontrib,
    Zhilin Zeng, Xiaokang Yang, Wei Shen\thanks{Corresponding Author.}
}
\affiliations{
MoE Key Lab of Artificial Intelligence, AI Institute, Shanghai Jiao Tong University \\
\{zelin.peng, fate311, bernardeschi, xkyang, wei.shen\}@sjtu.edu.cn

}

\usepackage{bibentry}

\begin{document}

\maketitle

\begin{abstract}

Segment Anything Model (SAM) has received remarkable attention as it offers a powerful and versatile solution for object segmentation in images. However, fine-tuning SAM for downstream segmentation tasks under different scenarios remains a challenge, as the varied characteristics of different scenarios naturally requires diverse model parameter spaces. Most existing fine-tuning methods attempt to bridge the gaps among different scenarios by introducing a set of new parameters to modify SAM's original parameter space. Unlike these works, in this paper, we propose fine-tuning SAM efficiently by \underline{\textbf{par}}ameter \underline{\textbf{s}}pac\underline{\textbf{e}} \underline{\textbf{r}}econstruction (SAM-PARSER), which introduce nearly zero trainable parameters during fine-tuning. In SAM-PARSER, we assume that SAM's original parameter space is relatively complete, so that its bases are able to reconstruct the parameter space of a new scenario. We obtain the bases by matrix decomposition, and fine-tuning the coefficients to reconstruct the parameter space tailored to the new scenario by an optimal linear combination of the bases. Experimental results show that SAM-PARSER exhibits superior segmentation performance across various scenarios, while reducing the number of trainable parameters by $\approx 290$ times compared with current parameter-efficient fine-tuning methods.

\end{abstract}

\section{Introduction}
\label{sec.1}

The recent unveiling of foundation models~\cite{GPT_nips_2020,SAM_arxiv_2023,seggpt_arxiv_2023} has shown unprecedented performance and potential across various domains in artificial intelligence. Among these, Segment Anything Model (SAM)~\cite{SAM_arxiv_2023} is one of the most noteworthy foundation models in computer vision, which contains a vast number of parameters and is pre-trained on a large-scale segmentation dataset, i.e., SA-1B. Consequently, SAM exhibits its capability in precise object segmentation across various images~\cite{zs_medical_arxiv_2023,zs_con_arxiv_2023}. The effectiveness of SAM has generated significant interest in fine-tuning it for downstream scenarios~\cite{SAMcustom_arxiv_2023,MedSAM_arxiv_2023}. However, this is a challenging task because the diverse characteristics inherent in different scenarios often necessitate varied model parameter spaces. Furthermore, given the vast number of parameters in SAM, direct fine-tuning of its original parameter space seems less feasible.

\begin{figure}[t]
  \centering
   \footnotesize
  \begin{overpic}[width=1.0\linewidth]{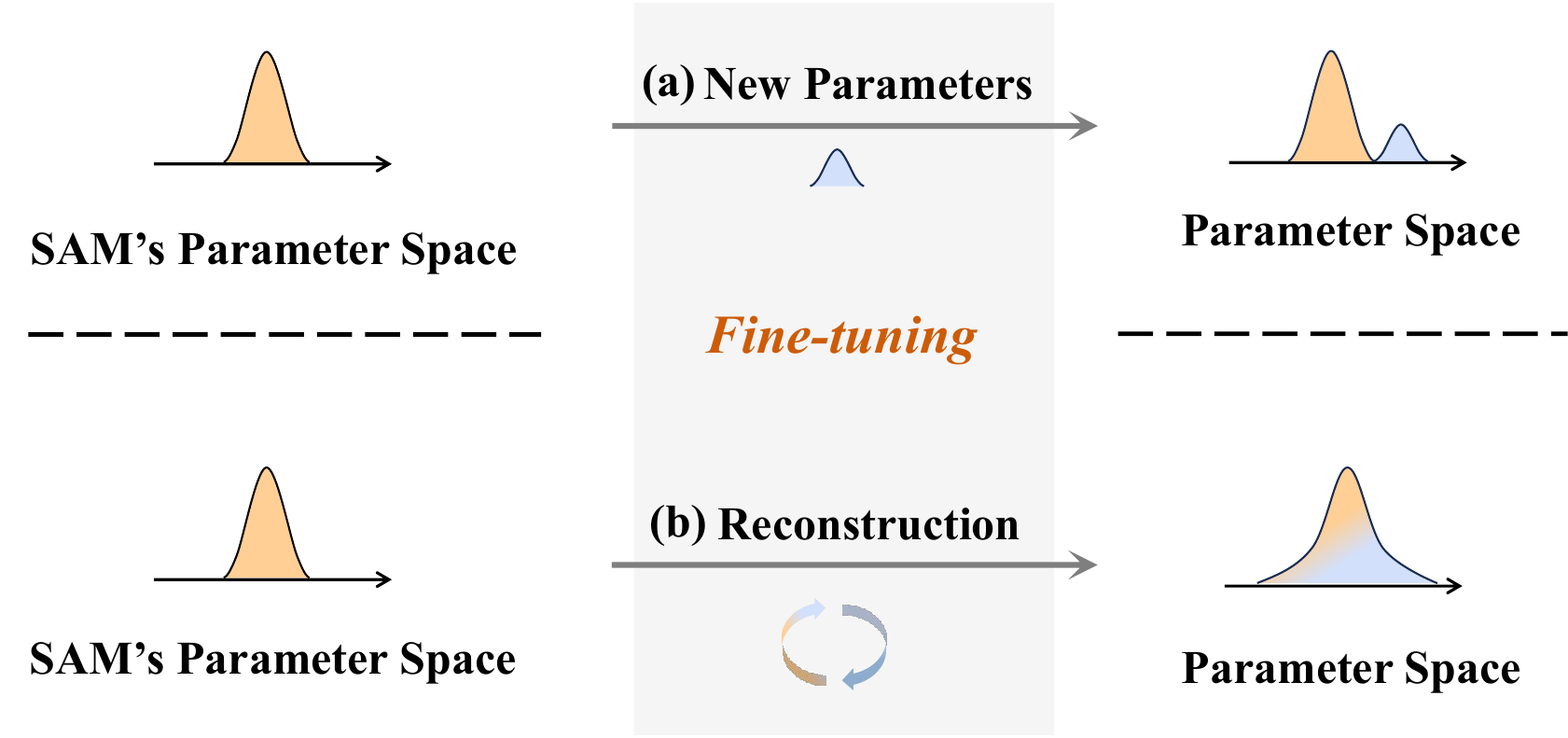}

  \end{overpic}
  \caption{\textbf{Comparative Overview: Previous Methods vs. SAM-PARSER.} (a) Previous parameter-efficient fine-tuning methods adapt SAM to different scenarios via adding new parameters. (b) Our SAM-PARSER adapts SAM to various scenarios through parameter space reconstruction, with nearly zero trainable parameters introduced.}
  \label{fig:1}
\end{figure}

Recent methods tackle these challenges by adopting a parameter-efficient fine-tuning paradigm, inspired by the prominent fine-tuning strategies in Natural Language Processing (NLP), i.e., LoRA~\cite{LORA_ICLR_2022} and Adapter~\cite{Adapter_ICML_2019}. As shown in Fig.~\ref{fig:1}, their common idea is to learn the distinct characteristics by introducing a set of new trainable parameters, thereby shifting the original parameter space to satisfy various downstream scenarios, leading to competitive performance for adapting SAM to new scenarios~\cite{MedSAM_arxiv_2023,SAMadapter_arxiv_2023,SonarSAM_arxiv_2023}. Considering that the original parameter space of SAM is already very huge, we raise a question by the light of nature: Is there any way to adapt SAM to new varied scenarios without introducing new trainable parameters?

In this paper, to offer a solution to this question, we propose fine-tuning SAM efficiently by \underline{\textbf{par}}ameter \underline{\textbf{s}}pac\underline{\textbf{e}} \underline{\textbf{r}}econstruction (SAM-PARSER). Since the original parameter space of SAM is huge, we can assume that its bases are foundational, which are capable of reconstructing the parameter space tailored to varied downstream scenarios. Specifically, we utilize a widely-accepted matrix decomposition technique, i.e., Singular Value Decomposition (SVD)~\cite{SVD_tc_1976}, decompose the original parameter space of SAM into the bases and associated coefficients. Then, by fine-tuning these coefficients, we are able to reconstruction of the parameter space tailored for new scenarios through an optimal linear combination of these bases. Consequently, our proposed SAM-PARSER efficiently adapts SAM to different scenarios with nearly zero newly introduced trainable parameters, i.e., a few coefficients, during fine-tuning.

Extensive experiments demonstrate that SAM-PARSER shows competitive performance across three prevalent scenarios in computer vision, including natural image segmentation, remote sensing image segmentation, and medical image segmentation. Additionally, different from many fine-tuning methods that target on the parameters of Transformer layers, we apply our SAM-PARSER solely to the convolutional layers in SAM's encoder. In such a situation, SAM-PARSER only needs to fine-tune 0.5k coefficients, reducing the number of trainable parameters by $\approx 290 \times$ compared with current parameter-efficient fine-tuning methods, while achieving superior segmentation performance.

\section{Related Work}

\subsection{Foundation Models}

Foundation models, initially introduced in the Natural Language Processing (NLP) community~\cite{NLP_foundation_arxiv_2021, GPT_nips_2020}, have marked a significant milestone in the trajectory of artificial intelligence. Among them, GPT~\cite{GPT_nips_2020}, with its advanced large language model, has drawn a lot of attention as its strong zero-shot generalization to unseen tasks and data. Recent developments have expanded the application of foundation models into the field of computer vision. Numerous vision foundation models~\cite{SAM_arxiv_2023,seggpt_arxiv_2023,SEEM_arxiv_2023} are developed to deal with plenty of vision tasks and data ditributions. Our experiments here are carried out using SAM~\cite{SAM_arxiv_2023}, a large foundation model trained on a large visual dataset, i.e., SA-1B. Large training data equips SAM with the capability to extract rich semantic features and detailed visual patterns, thereby ensuring its great potential for generalization across a variety of downstream scenarios.

\subsection{Fine-tuning SAM}

SAM consists of an image encoder and an image decoder, where the decoder is much more lightweight than the encoder. Thus, fully fine-tuning the decoder is a default operation in all SAM fine-tuning paradigms. The difference among different SAM fine-tuning paradigms lies in how to fine-tuning the encoder. For simplicity, we state \textbf{``fine-tuning SAM'' means ``fine-tuning SAM's image encoder''} in this paper. Naturally, a fine-tuning baseline is fully fine-tuning the decoder and fixing the encoder~\cite{MedSAM_arxiv_2023,only_decoder_shadow_arxiv_2023}.  
Then the fine-tuning paradigms for SAM can broadly be organized into two categories: 1) Fully fine-tuning, in which all SAM's parameters are fine-tuned; 2) Parameter-efficient fine-tuning, which freezes SAM's original parameter spaces and focuses on fine-tuning a small number of newly introduced parameters, e.g., LoRA~\cite{LORA_ICLR_2022}.

\noindent\textbf{Fully Fine-tuning.} A straightforward strategy is to fully fine-tune the entire parameters of SAM~\cite{SAM_arxiv_2023}, as highlighted in a previous study~\cite{POLY_SAM_arxiv_2023}. However, this approach inevitably demands substantial computational resources, which is less practical.

\noindent\textbf{Parameter-efficient Fine-tuning.}  These works~\cite{SonarSAM_arxiv_2023,SAMadapter_arxiv_2023,SAMadapter_arxiv_2023_2,SAMcustom_arxiv_2023,BadSAM_arxiv_2023,PE_SAM_arxiv_2023,HOW_SAM_arxiv_2023} sought to leverage insights derived from parameter-efficient fine-tuning paradigms in Natural Language Processing (NLP), such as Adapter~\cite{Adapter_ICML_2019} and LoRA~\cite{LORA_ICLR_2022}. 
For example, SAMed~\cite{SAMcustom_arxiv_2023} proposed the use of LoRA~\cite{LORA_ICLR_2022} to fine-tune SAM by optimizing the introduced rank decomposition matrices integrated into its Transformer blocks. Similarly, SAM-adapter~\cite{SAMadapter_arxiv_2023_2} employed the adapter technique~\cite{Adapter_ICML_2019} to fine-tune SAM. They introduced several trainable adapter layers in each Transformer layer of SAM, while concurrently freezing SAM's original parameter space. Contrary to these methods that adapt SAM to different scenarios by introducing new parameter spaces, our SAM-PARSER directly reconstructs SAM's original parameter space, offering a more straightforward and efficient solution for its fine-tuning.

\subsection{Parameter Space Decomposition}
Parameter space decomposition is a highly prevalent strategy for analyzing the structure of parameter spaces, broadly categorized into two main methods: matrix theory-based \cite{PENN_TNNLS_2021, EEPS_PRD_2011} and Fast Fourier Transform (FFT) theory-based methods~\cite{FNO_ICLR_2022}. Matrix theory-based decomposition typically involves analyzing components of the parameter space by techniques like Singular Value Decomposition (SVD)~\cite{PENN_TNNLS_2021} or QR decomposition~\cite{QRPS_NIPS_2021}. A prominent extension of this decomposition method is tensor theory-based decomposition~\cite{TD_ASA_2022}, commonly viewed as its higher-order counterpart. For the analysis of tensors, or other high-dimensional matrices, Tucker decomposition~\cite{FACT_AAAI_2023} is frequently employed. Differently, FFT theory-based decomposition typically entails mapping network weights into the frequency domain to analyze contributions across different frequency components~\cite{FT_arXiv_2023}. In this paper, we utilize matrix theory-based decomposition in our SAM-PARSER framework to reconstruct SAM's original parameter space, consistent with established practices in the contemporary parameter-efficient fine-tuning paradigm.

\section{Preliminaries}

In this section, we first look more closely at prevalent fine-tuning paradigms. Subsequently, we introduce the architecture of SAM. Ultimately, we discuss why parameter-efficient fine-tuning is favored for SAM and suggest a new perspective for its adaptation.

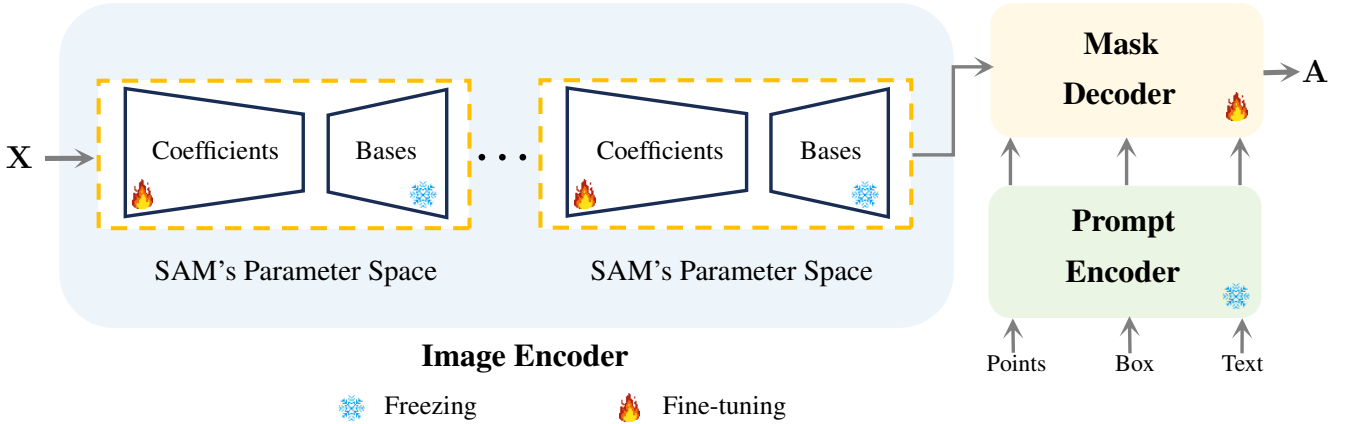
\begin{figure*}[t]
  \centering
  \small
  \begin{overpic}[width=0.95\linewidth]{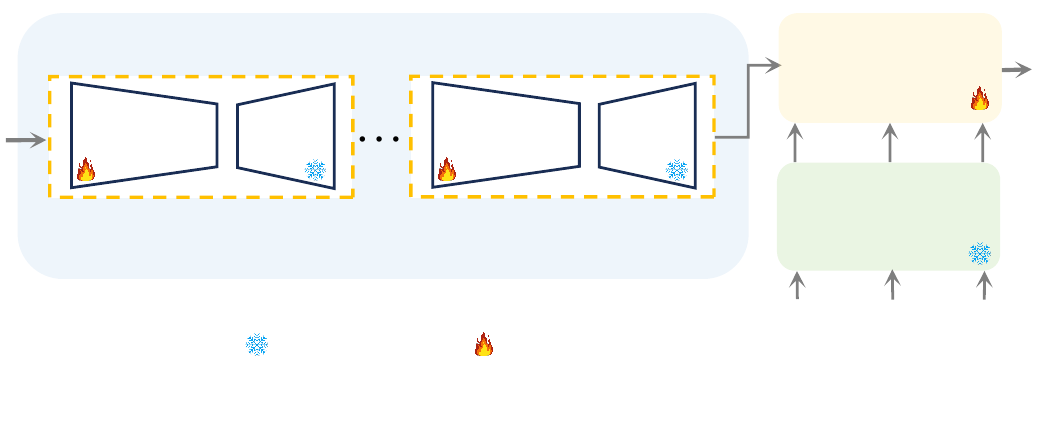}
  \put(27.3,8.6){\normalsize{Freezing}}
  \put(49.2,8.6){\normalsize{Fine-tuning}}

  \put(30.2,12.4){\Large{\textbf{Image Encoder}}}

  \put(-2.5,28){\Large{$\mathbf{X}$}}

  \put(99.5,34.8){\Large{$\mathbf{A}$}}

  \put(82.3,37){\Large{\textbf{Mask}}}
  \put(80.7,33){\Large{\textbf{Decoder}}}
  \put(81.4,23){\Large{\textbf{Prompt}}}
  \put(80.9,19){\Large{\textbf{Encoder}}}

  \put(74.7,12.0){Points}
  \put(84.8,12.0){Box}
  \put(93.2,12.0){Text}

  \put(9.2,19.2){\large{SAM's Parameter Space}}
  \put(43.5,19.2){\large{SAM's Parameter Space}}

  \put(9,28.7){\normalsize{Coefficients}}
  \put(25,28.7){\normalsize{Bases}}

  \put(44,28.7){\normalsize{Coefficients}}
  \put(60,28.7){\normalsize{Bases}}

  \end{overpic}
  \vspace{-37pt}
  \caption{\textbf{A schematic representation of SAM-PARSER.} We reconstruct SAM's parameter spaces for adapting to various scenarios by fine-tuning the coefficients while maintaining the foundational bases are frozen. Concurrently, we fully fine-tuning SAM's mask decoder.}

  \label{fig:global}
\end{figure*}

\subsection{Fine-tuning Paradigm}

\noindent\textbf{Fully Fine-tuning.}  Given a training set $\mathcal{T}$ used for fine-tuning, existing methods train a neural network $\mathcal{F}(\mathbf{X}; \textbf{W})$ to produce a dense prediction map $\mathbf{A}$, where $\mathbf{X}$ is an input image and $\textbf{W}$ are the trainable parameters of $\mathcal{F}$. Accordingly, the objective function of fully fine-tuning is formulated as:
\begin{equation}
    \textbf{W}^\ast=\arg\min_{\textbf{W}}\ell(\mathbf{A}, {\mathbf{Y}}),
\end{equation}
where $\mathbf{Y}$ is a full dense label map, and $\ell$ is typically chosen as a cross-entropy loss when addressing a segmentation task.

\noindent\textbf{Parameter-efficient Fine-tuning.} We provide a brief introduction of two representative parameter-efficient fine-tuning methods in NLP, i.e., Adapter~\cite{Adapter_ICML_2019} and LoRA~\cite{LORA_ICLR_2022}. 

\textbf{(1) Adapter.} Adapter~\cite{Adapter_ICML_2019} maintained all the original parameters frozen and integrated a learnable adapter layer between the Multi-Head Attention (MHA) module and the LayerNorm layer in each Transformer layer. This adapter layer, composed of two linear layers and a ReLU activation function, creates an extra parameter space specifically tailored to new scenarios. 

\textbf{(2) LoRA.} A prior work~\cite{NLP_low_dimension_arxiV_2020} illustrated that pre-trained large foundation models often reside on a low intrinsic dimension, allowing them to be projected into a smaller sub-space. Following this, LoRA~\cite{LORA_ICLR_2022} assumed that the change in weights during fine-tuning also has a low intrinsic rank. They introduced a learnable low-rank matrix $\Delta \textbf{W}$ that works in parallel to an original weighted matrix $\textbf{W} \in \mathbb{R}^{d \times k}$, which is in the MHA module of each transformer layer. $\Delta \textbf{W}$ is formed through a QR decomposition, denoted by $\Delta \textbf{W} = \bm{\beta}\bm{\alpha}$, where $\bm{\beta} \in \mathbb{R}^{d \times r}$, $\bm{\alpha} \in \mathbb{R}^{r \times k}$, and rank $r \ll \min(d, k)$. The size of $r$ determines the size of the newly introduced parameter space, allowing for a more targeted extraction of unique characteristics in new scenarios.

\subsection{Segment Anything Model (SAM)}    

SAM~\cite{SAM_arxiv_2023} is constructed of an image encoder, with extensive parameters, and a subsequent decoder. The image encoder consists of sequentially connected layers, starting with several transformer layers followed by standard convolutional layers. By training on a large-scale dataset, i.e., SA-1B, this encoder creates a broad parameter space that guides the decoder to achieve precise segmentation results for all objects contained in images. Additionally, to enable user interactions and identify specific objects, SAM is further equipped with a prompt encoder. This specialized encode is proficient in processing both dense (i.e., mask-based) and (i.e., box or point-based) prompts.

\subsection{Discussion}

Compared with fully fine-tuning paradigms, a parameter-efficient fine-tuning paradigm has attracted more attention in the fine-tuning SAM literature~\cite{SAMadapter_arxiv_2023,SonarSAM_arxiv_2023,POLY_SAM_arxiv_2023}. One reason is, as its name suggests, the parameter-efficient fine-tuning paradigm significantly reduces the trainable parameters comparing those of the fully fine-tuning paradigm, thus speeding up the training process. Besides, it is capable of learning unique characteristics of new scenarios by introducing new parameter spaces, making SAM better adapt to varied segmentation tasks.

In analyzing current fine-tuning literature, we find there is a clear trend toward reducing fine-tuning parameters while simultaneously aiming for improved results. Given this trajectory, we question if there exists a way that can match or even surpass the performance of prior paradigms with nearly zero newly introduced trainable parameters.

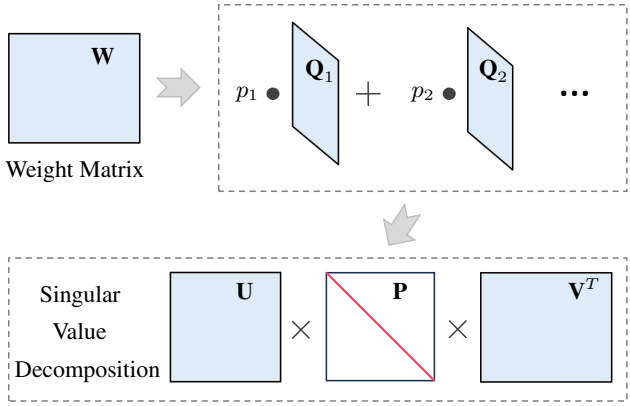
\begin{figure}[t]
  \centering
  \small
  \begin{overpic}[width=0.99\linewidth]{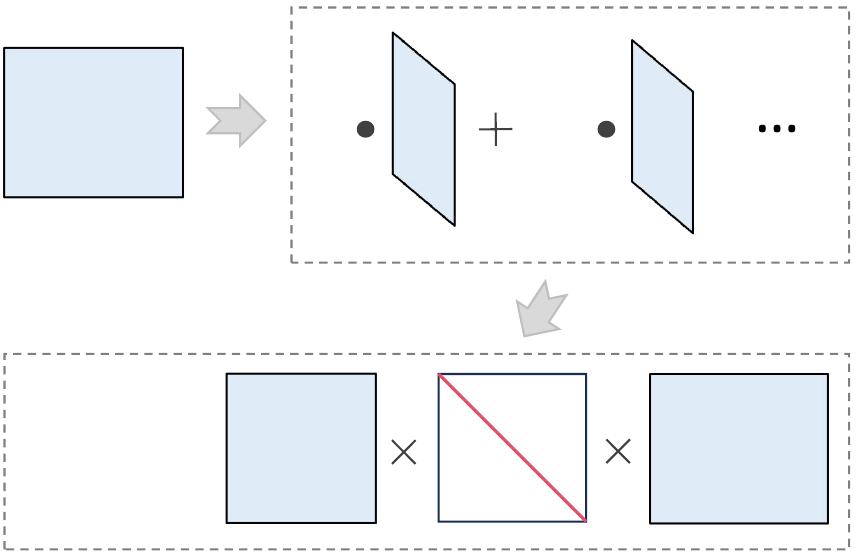}
   \put(5.5,16.3){Singular}   
   \put(7.5,10.3){Value} 
   \put(0.0,36.3){Weight Matrix} 

  \put(1.5,4.3){Decomposition}
  \put(36.8,16.8){$\textbf{U}$}
    \put(36.8,49.0){$p_1$}
    \put(64.8,49.0){$p_2$}

  \put(61.8,16.8){$\textbf{P}$}
  \put(89.8,16.8){$\textbf{V}^{T}$}
   \put(13.6,54.3){$\textbf{W}$} 
   \put(75.6,52.3){$\textbf{Q}_2$} 
   \put(47.9,52.3){$\textbf{Q}_1$} 

  \end{overpic}
  \caption{A schematic diagram to illustrate how to decompose weight matrices in SAM's image encoder.  First, we decompose the weight matrix $\textbf{W}$ into bases and their corresponding coefficients. Following this, we employ singular value decomposition (SVD)~\cite{SVD_tc_1976} to further disintegrate the bases $\textbf{Q}$ into a pair of bases, i.e., $\textbf{U}$ and $\textbf{V}$. }

  \label{fig:3}

\end{figure}

\section{Methodology}

In this section, we introduce our solution that proposes fine-tuning SAM efficiently by \underline{\textbf{par}}ameter \underline{\textbf{s}}pac\underline{\textbf{e}} \underline{\textbf{r}}econstruction (SAM-PARSER). In SAM-PARSER, we assume the bases of SAM's original parameter space are foundational, and capable of reconstructing the parameter space adapted to diverse downstream scenarios.
The core of the proposed SAM-PARSER lies in fine-tuning the coefficients corresponding to these bases, aiming to achieve the optimal linear combination for reconstructing the parameter spaces tailored to new scenarios. Fig.~\ref{fig:global} provides a visual overview of our proposed SAM-PARSER.

Specifically, for a pre-trained weight matrix $\textbf{W} \in \mathbb{R}^{d \times k}$, where $d \leq k$, it can be decomposed into bases and their corresponding coefficients, which are formulated as:
\begin{equation}
\label{eq_matrix_decom}
    \textbf{W}  = \sum_{i=1}^{d} p_i \textbf{Q}_i,
\end{equation}
where $p_i \in \mathbb{R}$ and $\textbf{Q}_i \in \mathbb{R}^{d \times k}$ are $i^{\text{th}}$ coefficient and base, respectively.

Following previous works~\cite{svdiff_arxiv_2023,SVDfew_nips_2022}, as shown in Fig.~\ref{fig:3}, we use a widely-accepted matrix decomposition technique, i.e., Singular Value Decomposition (SVD)~\cite{SVD_tc_1976}, to derive the bases and their coefficients. In this way, we can rewrite Eq.~\ref{eq_matrix_decom} as:
\begin{equation}
\begin{aligned}    
    \textbf{W} &= \sum_{i=1}^{d} p_i \textbf{u}_i (\textbf{v}_i)^{T} \\
                 &=  \textbf{U}\textbf{P}\textbf{V}^{T},    
\end{aligned}
\end{equation}
where we denote that $\textbf{u}_i \in \mathbb{R}^{d}$ and $\textbf{v}_i \in \mathbb{R}^{k}$, and then $\textbf{U} = \left[\textbf{u}_1, \textbf{u}_2, \ldots, \textbf{u}_i,  \ldots, \textbf{u}_d\right] \in \mathbb{R}^{d \times d}$ and $\textbf{V} = \left[\textbf{v}_1, \textbf{v}_2, \ldots, \textbf{v}_i,  \ldots, \textbf{v}_d \right] \in \mathbb{R}^{k \times d}$ compose of the bases of $\textbf{W}$. In our proposed SAM-PARSER, we only fine-tune the diagonal matrix \[
\textbf{P} = \begin{bmatrix}
p_1 & 0 & \cdots & 0 \\
0 & p_2 & \cdots & 0 \\
\vdots & \vdots & \ddots & \vdots \\
0 & 0 & \cdots & p_d
\end{bmatrix},
\] which are the coefficients. Accordingly, the objective function of our proposed SAM-PARSER is formulated as:
\begin{equation}
    \textbf{P}^\ast=\arg\min_{\textbf{P}}\ell(\mathbf{A}, {\mathbf{Y}}),
\end{equation}
as $\textbf{P}$ is a diagonal matrix, the number of parameters equals its rank, since the parameter matrix $\textbf{W}$ is a full-rank matrix. Unlike previous methods that simply freeze $\textbf{W}$ and introduce new parameters, we directly reconstruct $\textbf{W}$ by fine-tuning $\textbf{P}$, offering a more direct and efficient way for SAM adaptation across diverse scenarios.

\noindent\textbf{Loss Function.}
As recommended in previous works~\cite{SonarSAM_arxiv_2023,MedSAM_arxiv_2023}, we incorporate a combination of binary cross-entropy, denoted as $\ell_{ce}$, and binary dice loss, represented by $\ell_{dice}$, to fine-tune SAM. Then, for multi-instance scenes, we also adopt a focal loss $\ell_{focal}$~\cite{focal_loss_cvpr_2017} for balancing the learning of different instances. Finally, the loss function is derived as:
\begin{equation}
\ell = (1-\lambda)\ell_{ce} + \lambda\ell_{focal} + \ell_{dice},
\end{equation}
where $\lambda = 0$ represents a single-instance scene, commonly found in medical image segmentation, e.g., lesion segmentation.

\section{Experiments}

In this section, we evaluate our method across a spectrum of downstream segmentation tasks. These tasks fall into three primary categories: (1) Medical image segmentation, (2) Natural image segmentation, and (3) Remote sensing image segmentation. We begin by introducing the datasets utilized, the corresponding evaluation metrics, and implementation details. Following this, we delve into an ablation study of our proposed SAM-PARSER. Finally, we benchmark SAM-PARSER against various other fine-tuning strategies.

\subsection{Experimental Setup}

\noindent\textbf{Datasets.} To validate the effectiveness of our proposed methods, we conduct experiments on fine-tuning SAM to five datasets.

\textbf{(1) CT Abdominal organ (AO)}~\cite{AdomenCT_1K_Tpami_2022}. Following~\cite{MedSAM_arxiv_2023}, we randomly split 80\% medical images of AO for fine-tuning and 20\% for testing. 

\textbf{(2) COCO2017 (COCO)}~\cite{MSCOCO_2014_ECCV}. We fine-tune SAM by using natural images on the training set and evaluate the models' performance on its validation set. 

\textbf{(3) PASCAL VOC2012 (VOC)}~\cite{voc_2010_ijcv}. We fine-tune SAM by using natural images on its training set and evaluate the models' performance on its validation set. 

\textbf{(4) NWPU VHR-10 (NWPU)}~\cite{NWPU1_2014_ISPRS,NWPU2_2016_ISPRS,NWPU3_2016_TGRS}. As recommended in~\cite{NWPU3_2016_TGRS}, we allocate 70\% remote sensing images of NWPU for fine-tuning and the remaining 30\% for evaluation. 

\textbf{(5) WHU
building extraction (WHU)}~\cite{WHU_2016_TGRS}. It is tailored for remote sensing image segmentation. We fine-tune SAM on the training set and evaluate its performance on the validation set.







\noindent\textbf{Evaluation Metrics.} In line with previous studies~\cite{MedSAM_arxiv_2023,SonarSAM_arxiv_2023}, we utilize the Dice Similarity Coefficient (DSC) for evaluating medical image segmentation. For both natural and remote sensing image segmentation, we adopt mean intersection-over-union (mIoU) and the F1 score (F1) as our evaluation metrics, as recommended by~\cite{HHA_F1_miou_2022_PR}.

\noindent\textbf{Implementation Details.} 
In all of our experiments, we employ the ViT-Base version of SAM~\cite{SAM_arxiv_2023} as our backbone, integrating a box prompt for its prompt encoder input. In line with MedSAM~\cite{MedSAM_arxiv_2023}, we apply a random perturbation to each bounding box, varying between 0 and 20 pixels. Our training employs the Adam optimizer~\cite{adam_arxiv_2014}. 
For medical image segmentation, the initial learning rate is set to $10^{-6}$, and the weight decay is $5 \times 10^{-4}$ with one image per mini-batch. The number of fine-tuning epochs is set to 25. For natural and remote sensing image segmentation, we follow SonarSAM~\cite{SonarSAM_arxiv_2023}, the initial learning rate is set to $1.5 \times 10^{-4}$, and the weight decay is $5 \times 10^{-5}$ with one image per mini-batch. The number of fine-tuning epochs is set to 10.

\begin{table}[t]
\tabcolsep=0.09cm

\small
\centering
\begin{tabular}{c|cc|cc}

 & Trans. la. & Conv. la. & mIoU (\%) & F1 (\%) \\ \midrule[1.2pt]
{\color{Gray} Baseline$^\dag$} & - & - &{\color{Gray}82.9}&{\color{Gray}90.0} \\ \midrule
\multirow{3}{*}{\makecell{Parameter \\ Space}} &\checkmark &   &83.1 \color{Highlight}($+$0.2) &90.2 \color{Highlight}($+$0.2) \\ 
&\checkmark &  \checkmark& 83.3 \color{Highlight}($+$0.4) &90.3 \color{Highlight}($+$0.3)\\
&  & \checkmark & \textbf{84.2} \textbf{\color{Highlight}($+$1.3)} &\textbf{90.9} \textbf{\color{Highlight}($+$0.9)}\\
\end{tabular}
\caption{\textbf{Ablation on selecting which parameter space for reconstruction}. ``Trans. la.'': Transformer layers. ``Conv. la.'': Convolutional layers.  ``Baseline$^\dag$'': only fine-tuning SAM's decoder.}
\label{tab:ablation}
\end{table}

\begin{table}[t]
\tabcolsep=0.08cm

\small
\centering
\begin{tabular}{c|cc|cc}

 & Bases & Coefficients & mIoU (\%)  & Params (K)\\ \midrule[1.2pt]
{\color{Gray} Baseline$^\dag$} & - & - &{\color{Gray}82.9} &  {\color{Gray}3963.2} \\ \midrule
\multirow{3}{*}{\makecell{Component}} &  & \checkmark & 84.2 \color{Highlight}($+$1.3) &3963.7 \textbf{\color{Highlight}($+$0.5)}\\
&\checkmark &   &84.9 \color{Highlight}($+$2.0) &4852.4 \color{Highlight}($+$888.7)\\ 
&\checkmark &  \checkmark& \textbf{85.3} \textbf{\color{Highlight}($+$2.4)} &4852.9 \color{Highlight}($+$889.2)\\

\end{tabular}
\caption{\textbf{Ablation on fine-tuning bases or coefficients}. ``Baseline$^\dag$'': only fine-tuning SAM's decoder. Fine-tuning both the bases and coefficients equates to a full fine-tuning strategy, representing an upper bound of our SAM-PARSER.}
\label{tab:ablation1}
\end{table}

\subsection{Ablation Study}
\label{ablation}

SAM's whole parameter space is huge, where trainable parameters are mainly from transformer layers and convolutional layers. Previous fine-tuning methods predominantly target on the Transformer layers to introduce new parameters. Here, we delve into the parameters sub-spaces formed by the parameters from both the transformer layers and the convolutional layers and conduct ablation studies to determine reconstructing which parameter space/sub-space of SAM offers better performance. We also conduct an ablation study to verify our choice to fine-tune the coefficients rather than bases in SAM-PARSER. All results are reported on the NWPU dataset.

\begin{table*}
\centering
\tabcolsep=0.04cm
\begin{tabular}{l|ccccccccc}

\toprule
\multirow{2}{*}{Method} & AO & \multicolumn{2}{c}{COCO} & \multicolumn{2}{c}{VOC} & \multicolumn{2}{c}{NWPU} & \multicolumn{2}{c}{WHU} \\
\cmidrule(r){2-2} \cmidrule(r){3-4} \cmidrule(r){5-6} \cmidrule(r){7-8} \cmidrule(r){9-10}

 & DSC (\%) & mIoU (\%) & F1 (\%) & mIoU (\%) & F1 (\%) & mIoU (\%) & F1 (\%) & mIoU (\%) & F1 (\%) \\
\midrule
{\color{Gray} Baseline} & {\color{Gray} 66.4 \scriptsize{$\pm$ 1.1}} & {\color{Gray} 67.2 \scriptsize{$\pm$ 0.2}} & {\color{Gray} 76.3 \scriptsize{$\pm$ 0.1}} & {\color{Gray} 72.3 \scriptsize{$\pm$ 0.4}} & {\color{Gray} 81.7 \scriptsize{$\pm$ 0.5}} & {\color{Gray} 71.6 \scriptsize{$\pm$ 0.6}} & {\color{Gray} 81.4 \scriptsize{$\pm$ 0.5}} & {\color{Gray} 66.7 \scriptsize{$\pm$ 0.2}} & {\color{Gray} 76.0 \scriptsize{$\pm$ 0.3}} \\

Baseline$^\dag$ & 89.7 \scriptsize{$\pm$ 1.2}  & 74.4 \scriptsize{$\pm$ 0.2} & 82.7 \scriptsize{$\pm$ 0.3} & 78.7 \scriptsize{$\pm$ 0.5} & 88.3 \scriptsize{$\pm$ 0.2}  & 82.9 \scriptsize{$\pm$ 0.6}  & 90.0 \scriptsize{$\pm$ 0.4}  & 81.1 \scriptsize{$\pm$ 0.2}  & 87.5 \scriptsize{$\pm$ 0.1}  \\ \midrule

FacT~\cite{FACT_AAAI_2023} & 89.6 \scriptsize{$\pm$ 0.8} & 74.8 \scriptsize{$\pm$ 0.3} & 82.8 \scriptsize{$\pm$ 0.3} & 79.1 \scriptsize{$\pm$ 0.3} & 88.4 \scriptsize{$\pm$ 0.2} & 83.4  \scriptsize{$\pm$ 0.1} & 90.1 \scriptsize{$\pm$ 0.1} & 81.4  \scriptsize{$\pm$ 0.2} & 87.6 \scriptsize{$\pm$ 0.3} \\

LoRA~\cite{LORA_ICLR_2022} & 90.1 \scriptsize{$\pm$ 0.6} & \textbf{75.3} \scriptsize{$\pm$ 0.1} & 83.1 \scriptsize{$\pm$ 0.1} & 79.5 \scriptsize{$\pm$ 0.0} & 89.0 \scriptsize{$\pm$ 0.2} & 83.9 \scriptsize{$\pm$ 0.1} & 90.7 \scriptsize{$\pm$ 0.2} & \textbf{81.9}  \scriptsize{$\pm$ 0.1} & 88.2 \scriptsize{$\pm$ 0.1} \\

Adapter~\cite{Adapter_ICML_2019}  & 89.4 \scriptsize{$\pm$ 0.8} & 74.9 \scriptsize{$\pm$ 0.1} & 83.1 \scriptsize{$\pm$ 0.1} & 78.9 \scriptsize{$\pm$ 0.0} & 88.6 \scriptsize{$\pm$ 0.0} & 82.0 \scriptsize{$\pm$ 0.2} & 89.6 \scriptsize{$\pm$ 0.5} & 81.0 \scriptsize{$\pm$ 0.3}  & 87.5 \scriptsize{$\pm$ 0.4} \\

SAM-PARSER (Ours) & \textbf{90.9} \scriptsize{$\pm$ 0.3} & 75.0 \scriptsize{$\pm$ 0.1} & \textbf{83.2} \scriptsize{$\pm$ 0.0} & \textbf{79.6} \scriptsize{$\pm$ 0.1} & \textbf{89.2} \scriptsize{$\pm$ 0.1}  & \textbf{84.2} \scriptsize{$\pm$ 0.2} & \textbf{90.9} \scriptsize{$\pm$ 0.1} & 81.8 \scriptsize{$\pm$ 0.1} & \textbf{88.4} \scriptsize{$\pm$ 0.1} \\
\bottomrule
\end{tabular}
\caption{\textbf{Segment anything model (SAM) fine-tuned on five datasets.} ``AO'': CT Abdominal organ \emph{test} set~\cite{AdomenCT_1K_Tpami_2022} for medical image segmentation. ``COCO'': COCO2017 \emph{val} set~\cite{MSCOCO_2014_ECCV} for natural image segmentation. ``VOC'': PASCAL VOC2012 \emph{val} set~\cite{voc_2010_ijcv} from natural image segmentation. ``NWPU'': NWPU VHR-10 \emph{val} set~\cite{NWPU1_2014_ISPRS,NWPU2_2016_ISPRS,NWPU3_2016_TGRS} for remote sensing image segmentation. ``WHU'': WHU building extraction \emph{val} set~\cite{WHU_2016_TGRS} for remote sensing image segmentation. ``Baseline'': without any form of fine-tuning. ``Baseline$^\dag$'': only fine-tuning SAM's decoder. These results are evaluated with three different seeds.}
\label{tab:global}
\end{table*}



\begin{table}[t]
\tabcolsep=0.075cm

\small
\centering
\begin{tabular}{l|ll}
\toprule 

Method& Params (K) & Time (Fps) \\ \midrule
{\color{Gray} Baseline$^\dag$}&{\color{Gray}3963.2}& {\color{Gray}6.0} \\ \midrule
FacT~\cite{FACT_AAAI_2023}& 3977.5 \color{Highlight}\scriptsize{($+$14.3)}&3.3 \color{Highlight}\scriptsize{($-$2.7)} \\ 
LoRA~\cite{LORA_ICLR_2022}&4080.2 \color{Highlight}\scriptsize{($+$144)}&3.4 \color{Highlight}\scriptsize{($-$2.6)} \\ 
Adapter~\cite{Adapter_ICML_2019}&4314.7 \color{Highlight}\scriptsize{($+$351.5)}&2.9 \color{Highlight}\scriptsize{($-$3.1)}\\
SAM-PARSER (Ours)&\textbf{3963.7} \color{Highlight}\scriptsize{($+$\textbf{0.5})}&\textbf{5.8} \color{Highlight}\scriptsize{($-$\textbf{0.2})} \\
\bottomrule 
\end{tabular}
\caption{\textbf{Computation efficiency analysis for different fine-tuning methods.} ``Baseline$^\dag$'': only fine-tuning SAM's decoder.}
\label{tab:complexity}
\end{table}

\begin{table}[t]
\tabcolsep=0.15cm

\small
\centering
\begin{tabular}{l|cc}
\toprule 

Method& mIoU (\%) & F1 (\%) \\ \midrule
{\color{Gray} Baseline}&{\color{Gray}71.4}&{\color{Gray}82.7} \\
Baseline$^\dag$&81.2&89.4 \\ \midrule
FacT~\cite{FACT_AAAI_2023}& 81.7 \color{Highlight}($+$0.5) & 89.7 \color{Highlight}($+$0.3) \\ 
LoRA~\cite{LORA_ICLR_2022}& \textbf{82.4} \textbf{\color{Highlight}($+$1.2)} &\textbf{90.1} \textbf{\color{Highlight}($+$0.7)} \\ 
Adapter~\cite{Adapter_ICML_2019}&81.8 \color{Highlight}($+$0.6) &89.9 \color{Highlight}($+$0.5)\\
SAM-PARSER (Ours)& 81.6 \color{Highlight}($+$0.4) &89.6 \color{Highlight}($+$0.2)\\
\bottomrule 
\end{tabular}
\caption{\textbf{Extended experiments. Fine-tuning SAM on the \emph{val} set of SSDD dataset~\cite{SSDD_dataset_RS_2021}.} ``Baseline'': without any form of fine-tuning. ``Baseline$^\dag$'': only fine-tuning SAM's decoder.}
\label{tab:extend}
\end{table}

\noindent\textbf{Select which parameter space for reconstruction?} SAM's encoder consists of sequentially connected layers, starting with several transformer layers followed by standard convolutional layers. In this study, we approach the reconstruction of SAM's original parameter space through the reconstruction of its parameter sub-spaces, i.e., the sub-space formed by the parameters from 1) the transformer layers, 2) convolutional layers, and 3) both of them. As shown in Table.~\ref{tab:ablation}, only using parameters from transformer layers or using those from layers of both two types to form the parameter space for reconstruction leads to obvious performance drops, 0.9\% mIoU and 0.7\% mIoU, respectively. This indicates that the convolutional layers represent a more informative and crucial parameter space for reconstruction. Notably, when performing our method on the parameter space formed by the parameters only from the convolutional layers, we observe that the required trainable parameter size is only 0.5k while leading to superior performance.

\noindent\textbf{Fine-tuning bases or coefficients?} We ablate this experiment to validate the influence of fine-tuning different components in the original parameter space of SAM, i.e., coefficients and bases. As shown in Table.~\ref{tab:ablation1},
when compared with the strategies of solely fine-tuning the bases or fully fine-tuning both coefficients and bases (the latter representing the upper bound of our SAM-PARSER), our SAM-PARSER achieves a dramatic reduction in trainable parameters by approximately 1700 times. Surprisingly, the performance drop is marginal at only 0.7\% and 1.1\%, respectively. This demonstrates that fine-tuning coefficients offers an efficient way for adapting SAM to new scenarios.

\subsection{Main Results}
In this section, we compare our approach against several prevailing fine-tuning techniques for SAM. These include: (1) Exclusively fully fine-tuning SAM's decoder, which is used as our baseline, (2) Leveraging LoRA~\cite{LORA_ICLR_2022} for fine-tuning SAM's image encoder and fully fine-tuning SAM's decoder, and (3) Utilizing Adapter~\cite{Adapter_ICML_2019} for fine-tuning SAM's image encoder and fully fine-tuning SAM's decoder.

\noindent\textbf{Quantitative Results.} The quantitative results for fine-tuning SAM are shown in Tab~\ref{tab:global}.

\textbf{(1) Natural Image Segmentation.} Table~\ref{tab:global} gives the evaluation results comparing to other fine-tuning methods in natural image segmentation. On PASCAL VOC2012 \emph{val} set~\cite{voc_2010_ijcv}, our proposed method outperforms Adapter~\cite{Adapter_ICML_2019} by 0.7\% and 0.6\% in terms of mIoU and F1, respectively. But on COCO2017 \emph{val} set~\cite{MSCOCO_2014_ECCV}, our proposed method is worse than LoRA~\cite{LORA_ICLR_2022} by 0.2\% mIoU, a negative case we observe. We infer that for large natural image datasets, there is a significant overlap between their parameter spaces and SAM's original parameter space. Therefore, the gains from parameter space reconstruction are limited.

\begin{figure}[h]
  \centering
  \begin{overpic}[width=0.9\linewidth]{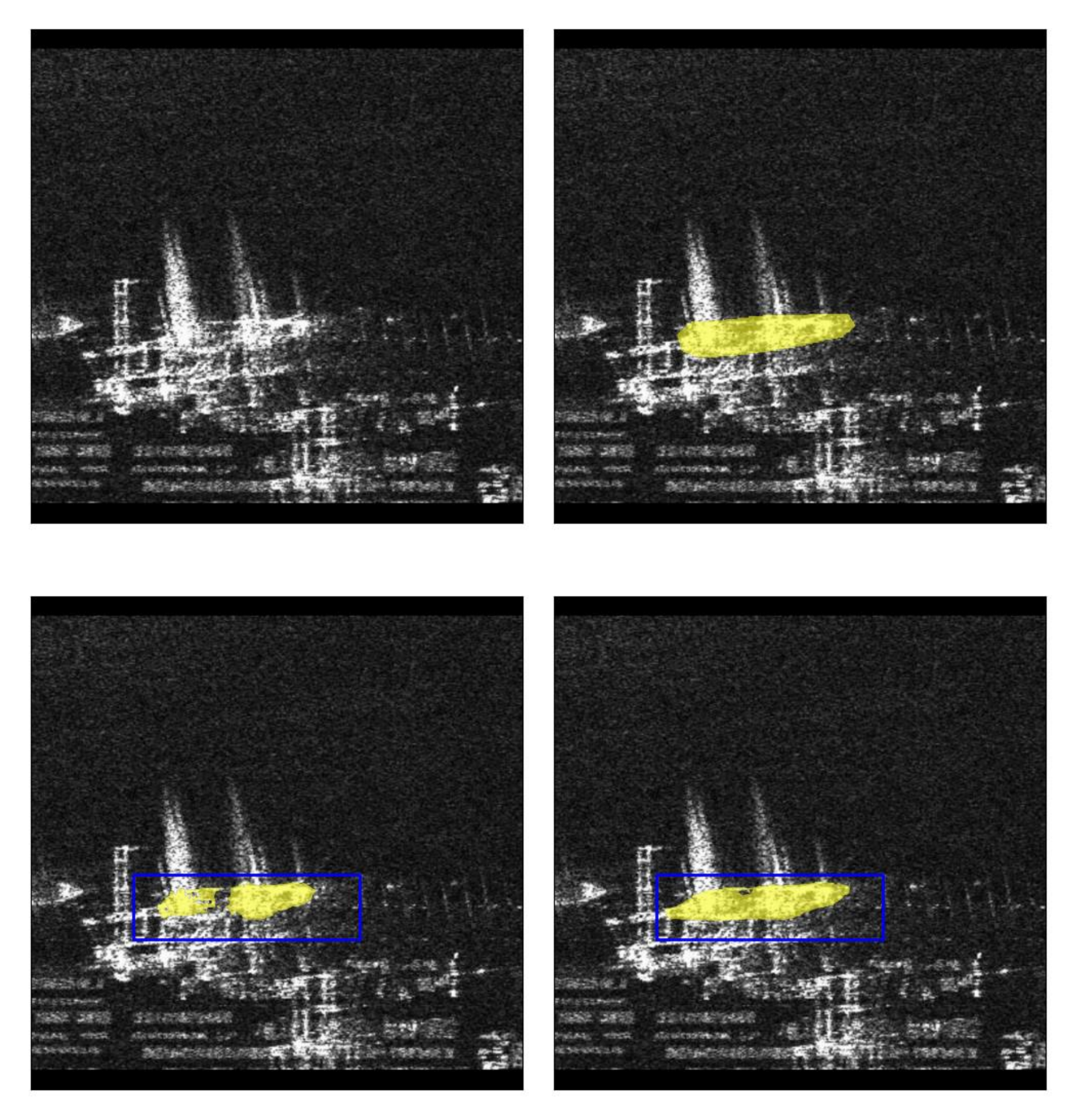}
  \put(17.0,48.4){\large{Input}}
  \put(61.0,48.4){\large{Ground-truth}}

  \put(17,-5){\large{Ours}}

  \put(61.0, -5){\large{LoRA}}

  \end{overpic}
  \vspace{10pt}
  \caption{\textbf{A failure case on SSDD \emph{val} set.}}

  \label{fig:rada}
  \vspace{-2pt}
\end{figure}

\begin{figure*}[t]
  \centering
  \begin{overpic}[width=0.99\textwidth]{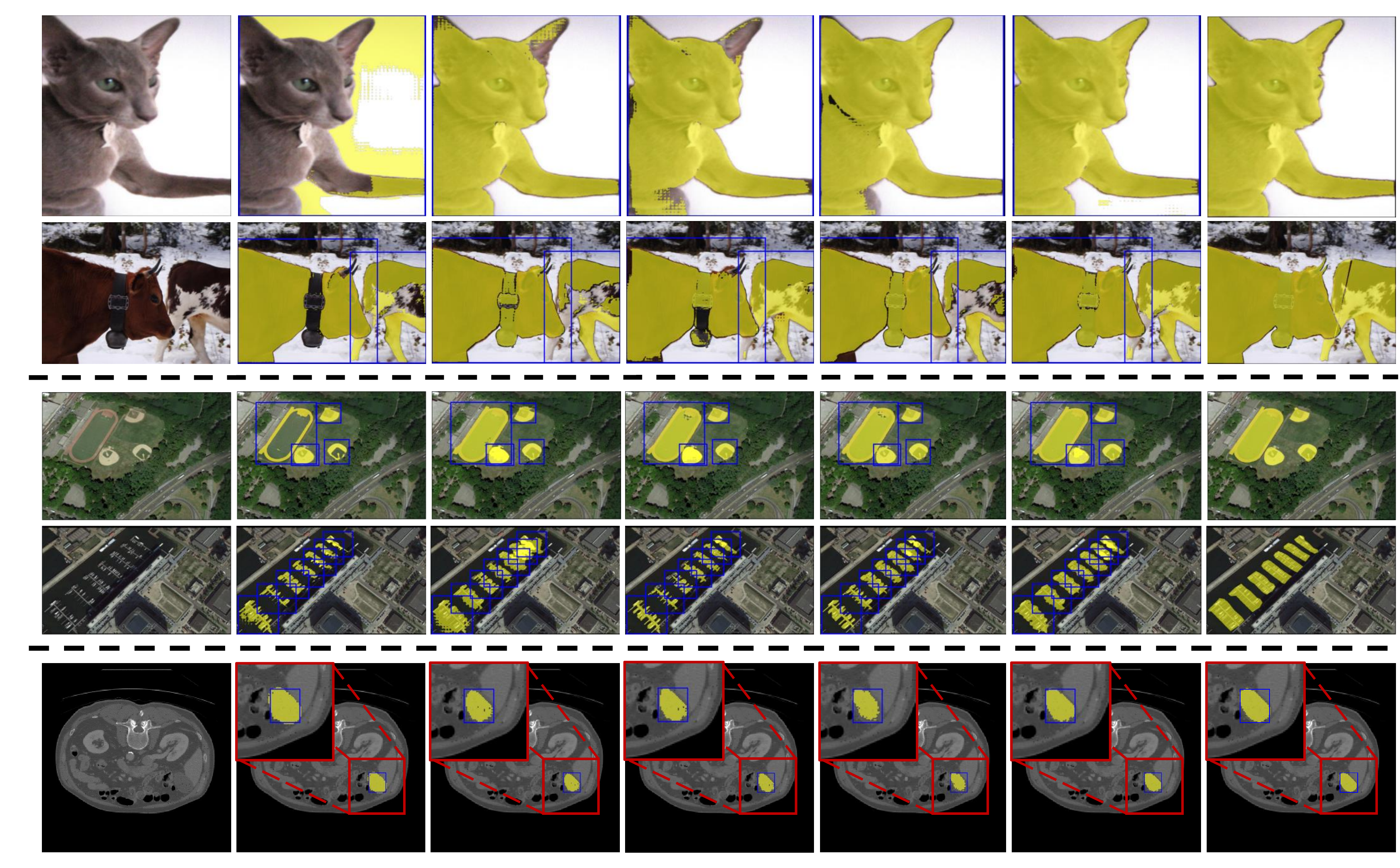}
  \put(0.0,45.5){\large{(a)}}
  \put(0.0,23.6){\large{(b)}}  
  \put(0.0,7){\large{(c)}}  

  \put(7,-2){\large{Input}}
  \put(19.8,-2){\large{Baseline}}
  \put(33.1,-2){\large{Baseline$^\dag$}}
  \put(48.6,-2){\large{LoRA}}
  \put(62,-2){\large{Adapter}}
  \put(77.5,-2){\large{Ours}}
  \put(87.5,-2){\large{Ground-truth}}

  \end{overpic}
  \vspace{12pt}
  \caption{\textbf{Qualitative segmentation results on three scenarios}, i.e., (a) natural image segmentation on PASCAL VOC2012 \emph{val} set~\cite{voc_2010_ijcv}, (b) remote sensing image segmentation on NWPU VHR-10 \emph{val} set~\cite{NWPU3_2016_TGRS}, and (c) medical image segmentation on CT Abdominal organ \emph{test} set. ``Baseline'': without any form of fine-tuning. ``Baseline$^\dag$'': only fine-tuning SAM's decoder.}

  \label{fig:vis}
\end{figure*}
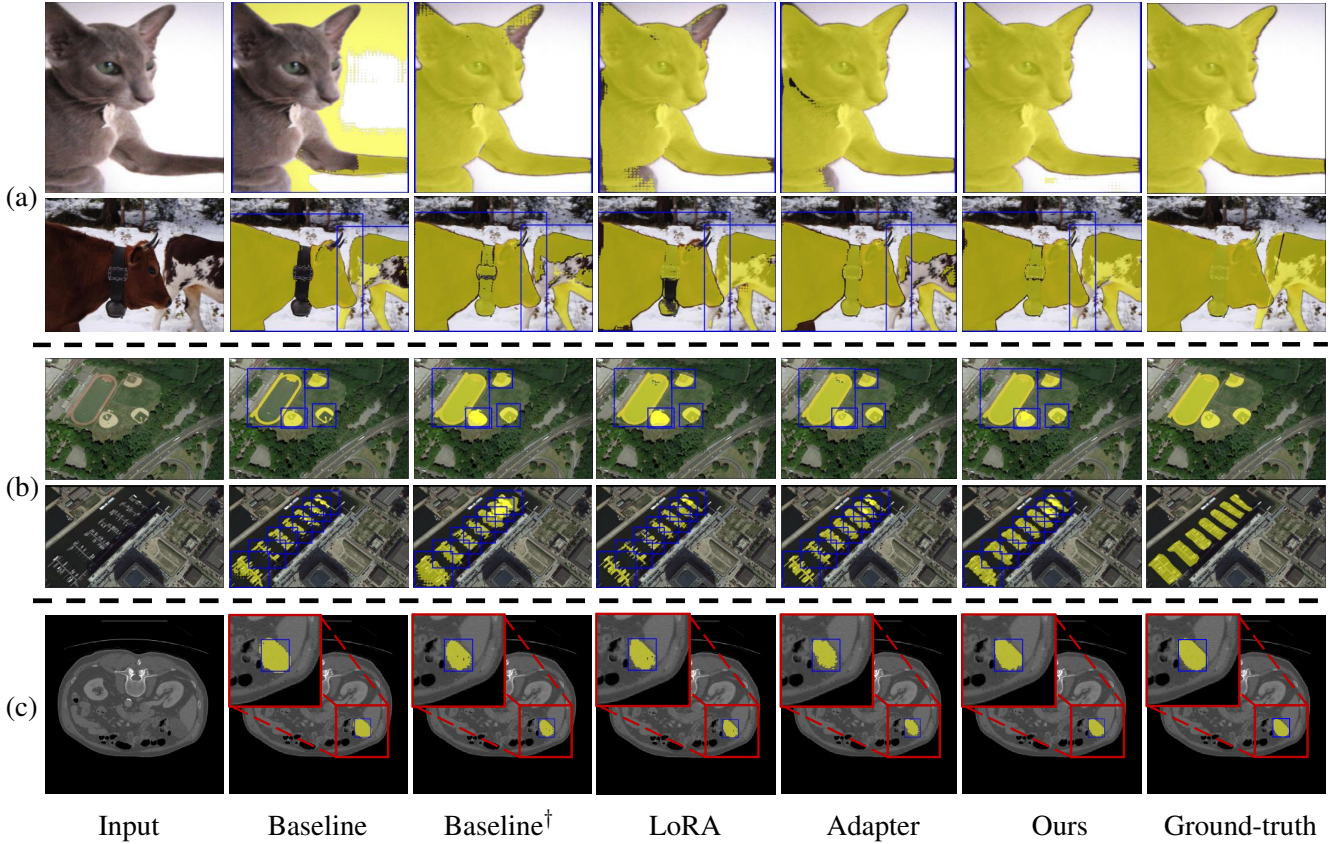

\textbf{(2) Remote Sensing Image Segmentation.} In Table~\ref{tab:global}, we compare our approach with others in remote sensing image segmentation. On both NWPU VHR-10 \emph{val} set and WHU building extraction \emph{val} set, our approach with little computing overhead boosts the baseline by a clear margin of 1.3\% mIoU and 0.8\% mIoU, respectively.

\textbf{(3) Medical Image Segmentation.} Table~\ref{tab:global} reports the comparison results on CT Abdominal organ test set~\cite{AdomenCT_1K_Tpami_2022}. It can be clearly seen that fine-tuning SAM with our proposed method outperforms the widely-used fine-tuning method LoRA~\cite{LORA_ICLR_2022} by 0.8\% DSC (from 90.3\% to 91.1\%). It demonstrates the effectiveness of our presented strategy for mitigating interfering information and aligning with the parameter space specific to medical image segmentation tasks.

\noindent\textbf{Qualitative results.} Here, we visualize some representative example segmentation results of our method against prevailing fine-tuning methods, e.g., LoRA~\cite{LORA_ICLR_2022} in three datasets. As shown in Fig~\ref{fig:vis}, we observe that our approach is able to diverse scenarios and produce more accurate results.

\noindent\textbf{Model Efficiency.} Our fine-tuning method brings performance improvements with nearly zero additional computational overhead. To validate this, we show the statistics of Params, i.e., the number of network trainable parameters, and Fps, i.e., training speed, in Table~\ref{tab:complexity}. It is clearly shown that the complexity of our method is significantly smaller than that of other fine-tuning methods. For example, the increase in Params in our method is \textbf{290} times less than that in LoRA~\cite{LORA_ICLR_2022}. This significant reduction in overhead stems from that our SAM-PARSER emphasizes reconstructing the coefficients in SAM's original parameter space, rather than creating an entirely new sub-space.

\noindent\textbf{Analysis of a failure case.} Our method is based on an assumption that the original parameter space of SAM is complete enough to cover new scenarios. However, this might be invalid when the image space of a new scenario is extremely different. To investigate this, we carry out extended experiments using the SSDD dataset~\cite{SSDD_dataset_RS_2021}, dedicated to radar image segmentation\textemdash \textbf{a stark departure from natural image segmentation.} As shown in Fig~\ref{fig:rada}, compared with LoRA, our SAM-PARSER achieves inferior segmentation results.
According to the experimental results in Table~\ref{tab:extend}, we can observe that our approach still outperforms the baseline by a mIoU of 0.4\%, but falls behind LoRA~\cite{LORA_ICLR_2022} and Adapter~\cite{Adapter_ICML_2019}. This demonstrates that in scenarios with distinctive characteristics, our proposed SAM-PARSER faces challenges as their parameter space cannot be reconstructed solely by the bases from the SAM's orignal parameter space.
As a potential solution, incorporating existing parameter-efficient methods, like LoRA, to expand the bases of the original parameter space, could further boost our method to reconstruct the parameter space of these scenarios.

\section{Conclusion}

We proposed fine-tuning SAM efficiently by parameter space reconstruction, called SAM-PARSER. In SAM-PARSER, we assume that SAM's initial parameter space is relatively complete, which allows us to use its bases for reconstruction of parameter space tailored to varied downstream scenarios. To achieve this, we employed SVD technique to decompose the original parameter space into the bases and associated coefficients. By fine-tuning these coefficients, we can achieve the optimal linear combination for reconstructing the parameter space of a new scenario. Extensive experiments on three different scenarios have demonstrated our superior performance, all while adding nearly zero trainable parameters.

\noindent\textbf{Acknowledgments.} This work was supported by NSFC 62322604, 62176159, Natural Science Foundation of Shanghai 21ZR1432200, and Shanghai Municipal Science and Technology Major Project 2021SHZDZX0102.


\bibliography{aaai24}

\begin{thebibliography}{41}
\providecommand{\natexlab}[1]{#1}

\bibitem[{Aghajanyan, Zettlemoyer, and Gupta(2020)}]{NLP_low_dimension_arxiV_2020}
Aghajanyan, A.; Zettlemoyer, L.; and Gupta, S. 2020.
\newblock Intrinsic dimensionality explains the effectiveness of language model fine-tuning.
\newblock \emph{arXiv preprint arXiv:2012.13255}.

\bibitem[{Andrews and Patterson(1976)}]{SVD_tc_1976}
Andrews, H.; and Patterson, C. 1976.
\newblock Singular value decomposition (SVD) image coding.
\newblock \emph{IEEE transactions on Communications}, 24(4): 425--432.

\bibitem[{Bommasani et~al.(2021)Bommasani, Hudson, Adeli, Altman, Arora, von Arx, Bernstein, Bohg, Bosselut, Brunskill et~al.}]{NLP_foundation_arxiv_2021}
Bommasani, R.; Hudson, D.~A.; Adeli, E.; Altman, R.; Arora, S.; von Arx, S.; Bernstein, M.~S.; Bohg, J.; Bosselut, A.; Brunskill, E.; et~al. 2021.
\newblock On the opportunities and risks of foundation models.
\newblock \emph{arXiv preprint arXiv:2108.07258}.

\bibitem[{Brown et~al.(2020)Brown, Mann, Ryder, Subbiah, Kaplan, Dhariwal, Neelakantan, Shyam, Sastry, Askell et~al.}]{GPT_nips_2020}
Brown, T.; Mann, B.; Ryder, N.; Subbiah, M.; Kaplan, J.~D.; Dhariwal, P.; Neelakantan, A.; Shyam, P.; Sastry, G.; Askell, A.; et~al. 2020.
\newblock Language models are few-shot learners.
\newblock \emph{Advances in neural information processing systems}, 33: 1877--1901.

\bibitem[{Cannon, Hanna, and Keppel(2011)}]{EEPS_PRD_2011}
Cannon, K.; Hanna, C.; and Keppel, D. 2011.
\newblock Efficiently enclosing the compact binary parameter space by singular-value decomposition.
\newblock \emph{Physical Review D}, 84(8): 084003.

\bibitem[{Chen et~al.(2023)Chen, Zhu, Ding, Cao, Zhang, Wang, Li, Sun, Mao, and Zang}]{SAMadapter_arxiv_2023}
Chen, T.; Zhu, L.; Ding, C.; Cao, R.; Zhang, S.; Wang, Y.; Li, Z.; Sun, L.; Mao, P.; and Zang, Y. 2023.
\newblock SAM Fails to Segment Anything?--SAM-Adapter: Adapting SAM in Underperformed Scenes: Camouflage, Shadow, and More.
\newblock \emph{arXiv preprint arXiv:2304.09148}.

\bibitem[{Cheng and Han(2016)}]{NWPU2_2016_ISPRS}
Cheng, G.; and Han, J. 2016.
\newblock A survey on object detection in optical remote sensing images.
\newblock \emph{ISPRS journal of photogrammetry and remote sensing}, 117: 11--28.

\bibitem[{Cheng et~al.(2014)Cheng, Han, Zhou, and Guo}]{NWPU1_2014_ISPRS}
Cheng, G.; Han, J.; Zhou, P.; and Guo, L. 2014.
\newblock Multi-class geospatial object detection and geographic image classification based on collection of part detectors.
\newblock \emph{ISPRS Journal of Photogrammetry and Remote Sensing}, 98: 119--132.

\bibitem[{Cheng, Zhou, and Han(2016)}]{NWPU3_2016_TGRS}
Cheng, G.; Zhou, P.; and Han, J. 2016.
\newblock Learning rotation-invariant convolutional neural networks for object detection in VHR optical remote sensing images.
\newblock \emph{IEEE Transactions on Geoscience and Remote Sensing}, 54(12): 7405--7415.

\bibitem[{Dutt et~al.(2023)Dutt, Ericsson, Sanchez, Tsaftaris, and Hospedales}]{PE_SAM_arxiv_2023}
Dutt, R.; Ericsson, L.; Sanchez, P.; Tsaftaris, S.~A.; and Hospedales, T. 2023.
\newblock Parameter-Efficient Fine-Tuning for Medical Image Analysis: The Missed Opportunity.
\newblock \emph{arXiv preprint arXiv:2305.08252}.

\bibitem[{Everingham et~al.(2010)Everingham, Van~Gool, Williams, Winn, and Zisserman}]{voc_2010_ijcv}
Everingham, M.; Van~Gool, L.; Williams, C.~K.; Winn, J.; and Zisserman, A. 2010.
\newblock The pascal visual object classes (voc) challenge.
\newblock \emph{International journal of computer vision}, 88: 303--338.

\bibitem[{Guan et~al.(2023)Guan, Hu, Zhou, Zhang, Li, and Liu}]{BadSAM_arxiv_2023}
Guan, Z.; Hu, M.; Zhou, Z.; Zhang, J.; Li, S.; and Liu, N. 2023.
\newblock Badsam: Exploring security vulnerabilities of sam via backdoor attacks.
\newblock \emph{arXiv preprint arXiv:2305.03289}.

\bibitem[{Han et~al.(2023)Han, Li, Zhang, Milanfar, Metaxas, and Yang}]{svdiff_arxiv_2023}
Han, L.; Li, Y.; Zhang, H.; Milanfar, P.; Metaxas, D.; and Yang, F. 2023.
\newblock Svdiff: Compact parameter space for diffusion fine-tuning.
\newblock \emph{arXiv preprint arXiv:2303.11305}.

\bibitem[{He et~al.(2023{\natexlab{a}})He, Bao, Li, Grant, and Ou}]{zs_medical_arxiv_2023}
He, S.; Bao, R.; Li, J.; Grant, P.~E.; and Ou, Y. 2023{\natexlab{a}}.
\newblock Accuracy of segment-anything model (sam) in medical image segmentation tasks.
\newblock \emph{arXiv preprint arXiv:2304.09324}.

\bibitem[{He et~al.(2023{\natexlab{b}})He, Yang, Feng, Yin, Wang, Leng, and Lin}]{FT_arXiv_2023}
He, Z.; Yang, M.; Feng, M.; Yin, J.; Wang, X.; Leng, J.; and Lin, Z. 2023{\natexlab{b}}.
\newblock Fourier Transformer: Fast Long Range Modeling by Removing Sequence Redundancy with FFT Operator.
\newblock \emph{arXiv preprint arXiv:2305.15099}.

\bibitem[{Houlsby et~al.(2019)Houlsby, Giurgiu, Jastrzebski, Morrone, De~Laroussilhe, Gesmundo, Attariyan, and Gelly}]{Adapter_ICML_2019}
Houlsby, N.; Giurgiu, A.; Jastrzebski, S.; Morrone, B.; De~Laroussilhe, Q.; Gesmundo, A.; Attariyan, M.; and Gelly, S. 2019.
\newblock Parameter-efficient transfer learning for NLP.
\newblock In \emph{International Conference on Machine Learning}, 2790--2799. PMLR.

\bibitem[{Hu et~al.(2022)Hu, Shen, Wallis, Allen-Zhu, Li, Wang, Wang, and Chen}]{LORA_ICLR_2022}
Hu, E.~J.; Shen, Y.; Wallis, P.; Allen-Zhu, Z.; Li, Y.; Wang, S.; Wang, L.; and Chen, W. 2022.
\newblock Lo{RA}: Low-Rank Adaptation of Large Language Models.
\newblock In \emph{International Conference on Learning Representations}.

\bibitem[{Ji et~al.(2023)Ji, Fan, Xu, Cheng, Zhou, and Van~Gool}]{zs_con_arxiv_2023}
Ji, G.-P.; Fan, D.-P.; Xu, P.; Cheng, M.-M.; Zhou, B.; and Van~Gool, L. 2023.
\newblock SAM Struggles in Concealed Scenes--Empirical Study on" Segment Anything".
\newblock \emph{arXiv preprint arXiv:2304.06022}.

\bibitem[{Ji, Wei, and Lu(2018)}]{WHU_2016_TGRS}
Ji, S.; Wei, S.; and Lu, M. 2018.
\newblock Fully convolutional networks for multisource building extraction from an open aerial and satellite imagery data set.
\newblock \emph{IEEE Transactions on geoscience and remote sensing}, 57(1): 574--586.

\bibitem[{Jie and Deng(2023)}]{FACT_AAAI_2023}
Jie, S.; and Deng, Z.-H. 2023.
\newblock Fact: Factor-tuning for lightweight adaptation on vision transformer.
\newblock In \emph{Proceedings of the AAAI Conference on Artificial Intelligence}, 1, 1060--1068.

\bibitem[{Kingma and Ba(2014)}]{adam_arxiv_2014}
Kingma, D.~P.; and Ba, J. 2014.
\newblock Adam: A method for stochastic optimization.
\newblock \emph{arXiv preprint arXiv:1412.6980}.

\bibitem[{Kirillov et~al.(2023)Kirillov, Mintun, Ravi, Mao, Rolland, Gustafson, Xiao, Whitehead, Berg, Lo, Doll{\'a}r, and Girshick}]{SAM_arxiv_2023}
Kirillov, A.; Mintun, E.; Ravi, N.; Mao, H.; Rolland, C.; Gustafson, L.; Xiao, T.; Whitehead, S.; Berg, A.~C.; Lo, W.-Y.; Doll{\'a}r, P.; and Girshick, R. 2023.
\newblock Segment Anything.
\newblock \emph{arXiv:2304.02643}.

\bibitem[{Li, Hu, and Yang(2023)}]{POLY_SAM_arxiv_2023}
Li, Y.; Hu, M.; and Yang, X. 2023.
\newblock Polyp-sam: Transfer sam for polyp segmentation.
\newblock \emph{arXiv preprint arXiv:2305.00293}.

\bibitem[{Li et~al.(2022)Li, Kovachki, Azizzadenesheli, Liu, Bhattacharya, Stuart, and Anandkumar}]{FNO_ICLR_2022}
Li, Z.; Kovachki, N.; Azizzadenesheli, K.; Liu, B.; Bhattacharya, K.; Stuart, A.; and Anandkumar, A. 2022.
\newblock Fourier neural operator for parametric partial differential equations.
\newblock In \emph{The Eleventh International Conference on Learning Representations}.

\bibitem[{Lin et~al.(2017)Lin, Goyal, Girshick, He, and Doll{\'a}r}]{focal_loss_cvpr_2017}
Lin, T.-Y.; Goyal, P.; Girshick, R.; He, K.; and Doll{\'a}r, P. 2017.
\newblock Focal loss for dense object detection.
\newblock In \emph{Proceedings of the IEEE international conference on computer vision}, 2980--2988.

\bibitem[{Lin et~al.(2014)Lin, Maire, Belongie, Hays, Perona, Ramanan, Doll{\'a}r, and Zitnick}]{MSCOCO_2014_ECCV}
Lin, T.-Y.; Maire, M.; Belongie, S.; Hays, J.; Perona, P.; Ramanan, D.; Doll{\'a}r, P.; and Zitnick, C.~L. 2014.
\newblock Microsoft coco: Common objects in context.
\newblock In \emph{Computer Vision--ECCV 2014: 13th European Conference, Zurich, Switzerland, September 6-12, 2014, Proceedings, Part V 13}, 740--755. Springer.

\bibitem[{Ma and Wang(2023)}]{MedSAM_arxiv_2023}
Ma, J.; and Wang, B. 2023.
\newblock Segment anything in medical images.
\newblock \emph{arXiv preprint arXiv:2304.12306}.

\bibitem[{Ma et~al.(2022)Ma, Zhang, Gu, Zhu, Ge, Zhang, An, Wang, Wang, Liu, Cao, Zhang, Liu, Wang, Li, He, and Yang}]{AdomenCT_1K_Tpami_2022}
Ma, J.; Zhang, Y.; Gu, S.; Zhu, C.; Ge, C.; Zhang, Y.; An, X.; Wang, C.; Wang, Q.; Liu, X.; Cao, S.; Zhang, Q.; Liu, S.; Wang, Y.; Li, Y.; He, J.; and Yang, X. 2022.
\newblock AbdomenCT-1K: Is Abdominal Organ Segmentation a Solved Problem?
\newblock \emph{IEEE Transactions on Pattern Analysis and Machine Intelligence}, 44(10): 6695--6714.

\bibitem[{Mijnsbrugge, Ongenae, and Van~Hoecke(2021)}]{PENN_TNNLS_2021}
Mijnsbrugge, D.~V.; Ongenae, F.; and Van~Hoecke, S. 2021.
\newblock Parameter Efficient Neural Networks With Singular Value Decomposed Kernels.
\newblock \emph{IEEE Transactions on Neural Networks and Learning Systems}, 1--11.

\bibitem[{Panahi, Saeedi, and Arodz(2021)}]{QRPS_NIPS_2021}
Panahi, A.; Saeedi, S.; and Arodz, T. 2021.
\newblock Shapeshifter: a parameter-efficient transformer using factorized reshaped matrices.
\newblock \emph{Advances in Neural Information Processing Systems}, 34: 1337--1350.

\bibitem[{Sun et~al.(2022)Sun, Chen, He, Wang, Feng, Han, Ding, Cheng, Li, and Wang}]{SVDfew_nips_2022}
Sun, Y.; Chen, Q.; He, X.; Wang, J.; Feng, H.; Han, J.; Ding, E.; Cheng, J.; Li, Z.; and Wang, J. 2022.
\newblock Singular value fine-tuning: Few-shot segmentation requires few-parameters fine-tuning.
\newblock \emph{Advances in Neural Information Processing Systems}, 35: 37484--37496.

\bibitem[{Wang et~al.(2022)Wang, Zheng, Lian, and Li}]{TD_ASA_2022}
Wang, D.; Zheng, Y.; Lian, H.; and Li, G. 2022.
\newblock High-dimensional vector autoregressive time series modeling via tensor decomposition.
\newblock \emph{Journal of the American Statistical Association}, 117(539): 1338--1356.

\bibitem[{Wang et~al.(2023{\natexlab{a}})Wang, Ye, Zhu, Wu, Zhang, Xing, and Hu}]{SonarSAM_arxiv_2023}
Wang, L.; Ye, X.; Zhu, L.; Wu, W.; Zhang, J.; Xing, H.; and Hu, C. 2023{\natexlab{a}}.
\newblock When SAM Meets Sonar Images.
\newblock \emph{arXiv preprint arXiv:2306.14109}.

\bibitem[{Wang et~al.(2023{\natexlab{b}})Wang, Zhang, Cao, Wang, Shen, and Huang}]{seggpt_arxiv_2023}
Wang, X.; Zhang, X.; Cao, Y.; Wang, W.; Shen, C.; and Huang, T. 2023{\natexlab{b}}.
\newblock Seggpt: Segmenting everything in context.
\newblock \emph{arXiv preprint arXiv:2304.03284}.

\bibitem[{Wang et~al.(2023{\natexlab{c}})Wang, Zhou, Mao, and Li}]{only_decoder_shadow_arxiv_2023}
Wang, Y.; Zhou, W.; Mao, Y.; and Li, H. 2023{\natexlab{c}}.
\newblock Detect Any Shadow: Segment Anything for Video Shadow Detection.
\newblock \emph{arXiv preprint arXiv:2305.16698}.

\bibitem[{Wu et~al.(2023)Wu, Fu, Fang, Liu, Wang, Xu, Jin, and Arbel}]{SAMadapter_arxiv_2023_2}
Wu, J.; Fu, R.; Fang, H.; Liu, Y.; Wang, Z.; Xu, Y.; Jin, Y.; and Arbel, T. 2023.
\newblock Medical sam adapter: Adapting segment anything model for medical image segmentation.
\newblock \emph{arXiv preprint arXiv:2304.12620}.

\bibitem[{Zhang and Liu(2023)}]{SAMcustom_arxiv_2023}
Zhang, K.; and Liu, D. 2023.
\newblock Customized segment anything model for medical image segmentation.
\newblock \emph{arXiv preprint arXiv:2304.13785}.

\bibitem[{Zhang et~al.(2021)Zhang, Zhang, Li, Xu, Wang, Zhan, Xu, Ke, Zeng, Su et~al.}]{SSDD_dataset_RS_2021}
Zhang, T.; Zhang, X.; Li, J.; Xu, X.; Wang, B.; Zhan, X.; Xu, Y.; Ke, X.; Zeng, T.; Su, H.; et~al. 2021.
\newblock SAR ship detection dataset (SSDD): Official release and comprehensive data analysis.
\newblock \emph{Remote Sensing}, 13(18): 3690.

\bibitem[{Zhang and Jiao(2023)}]{HOW_SAM_arxiv_2023}
Zhang, Y.; and Jiao, R. 2023.
\newblock How Segment Anything Model (SAM) Boost Medical Image Segmentation?
\newblock \emph{arXiv preprint arXiv:2305.03678}.

\bibitem[{Zheng et~al.(2022)Zheng, Gong, Liu, Jiang, Zhan, Lu, and Zhang}]{HHA_F1_miou_2022_PR}
Zheng, H.; Gong, M.; Liu, T.; Jiang, F.; Zhan, T.; Lu, D.; and Zhang, M. 2022.
\newblock HFA-Net: High frequency attention siamese network for building change detection in VHR remote sensing images.
\newblock \emph{Pattern Recognition}, 129: 108717.

\bibitem[{Zou et~al.(2023)Zou, Yang, Zhang, Li, Li, Gao, and Lee}]{SEEM_arxiv_2023}
Zou, X.; Yang, J.; Zhang, H.; Li, F.; Li, L.; Gao, J.; and Lee, Y.~J. 2023.
\newblock Segment everything everywhere all at once.
\newblock \emph{arXiv preprint arXiv:2304.06718}.

\end{thebibliography}

\end{document}